\documentclass[runningheads]{llncs}

\usepackage{accv}

\usepackage{accvabbrv}
\usepackage{multirow}
\usepackage{multicol}
\usepackage{graphicx}%
\usepackage{amsmath,amssymb,amsfonts}%
\usepackage{csquotes}
\usepackage{mathrsfs}%
\usepackage[title]{appendix}%
\usepackage{textcomp}%
\usepackage{manyfoot}%
\usepackage{booktabs}%
\usepackage{algorithm}%
\usepackage{algorithmicx}%
\usepackage{algpseudocode}%
\usepackage{listings}%
\usepackage[table]{xcolor}
\usepackage[export]{adjustbox}

\usepackage{graphicx}
\usepackage{booktabs}

\usepackage[accsupp]{axessibility}  

\usepackage[pagebackref,breaklinks,colorlinks,citecolor=accvblue]{hyperref}

\usepackage{orcidlink}

\begin{document}

\title{TDiR: Transformer based Diffusion for Image Restoration Tasks} 

\titlerunning{~}

\author{Abbas Anwar\inst{1,2}\and
Ibrahim Radwan\inst{3}\and
Ali Arshad Nasir\inst{4}\and \\
Mudassir Masood\inst{4}\and
Saeed Anwar\inst{5}
}

\authorrunning{A. Anwar~et~al.}
\institute{University of Technology Nowshera, Pakistan \\ \and
National University of Computer and Emerging Sciences, Pakistan\\ \and
The University of Canberra, Australia\\ \and
King Fahd University of Petroleum and Minerals, KSA\\ \and
The University of Western Australia, Australia\\
}

\maketitle

\begin{abstract}
Images captured in challenging environments often experience various types of degradation, such as noise, color cast, blur, and light scattering. These issues significantly lower image quality, thereby reducing their usefulness in downstream tasks such as object detection, mapping, and classification. Our transformer-based diffusion model was developed to address image restoration challenges and enhance the quality of degraded images. Our methodology is assessed across three primary image restoration tasks, including underwater enhancement, denoising, and deraining, utilizing five standard benchmarks. It is then compared to 18 state-of-the-art techniques, employing four evaluation metrics. Our results show that the diffusion model, combined with transformers, outperforms current methods. The findings highlight the effectiveness of diffusion models and transformers in improving degraded image quality, thereby broadening their application in downstream tasks that demand high-fidelity visual data.
\end{abstract}

\section{Introduction}

Image restoration is a fundamental problem in computer vision and image processing, aiming to recover an original, clean image from a corrupted or degraded version~\cite{anwar2015class}. Corruption can arise from various sources, such as noise, blur, and compression artifacts. Image restoration improves image quality across diverse applications, including medical imaging, satellite imaging, surveillance, and photography. Restoration techniques~\cite{Anwar2021R2NET} enable better interpretation, analysis, and decision-making across various fields by reconstructing compromised images.

In medical imaging~\cite{Qian_2026_CVPR}, image precision and clarity are paramount for accurate diagnosis and treatment planning. Noise or artifacts in medical images may obscure essential details, potentially leading to misdiagnoses or missed conditions. Likewise, restoring clear, precise images in satellite imagery~\cite{Daroya_2025_ICCV} is crucial for monitoring environmental changes, managing natural disasters, and supporting agricultural activities. High-quality, restored images provide more reliable data for researchers and policymakers, thereby enabling informed decision-making grounded in accurate visual information.

Furthermore, in the domains of surveillance and security~\cite{zhang2026out}, image restoration significantly enhances the ability to identify and analyze objects, faces, and activities captured by cameras, particularly under low-light or adverse-weather conditions~\cite{Wang_2025_ICCV}. This capability is vital for ensuring public safety and effective law enforcement. Within the realm of photography, both amateur and professional photographers benefit from image restoration techniques that can salvage photographs compromised by motion blur or inadequate lighting, thereby preserving valuable memories and artistic works. Overall, advances in image restoration methodologies continue to broaden the potential and reliability of visual data across a wide spectrum of critical applications.

In this study, we implement a transformer-based U-Net architecture~\cite{Tian2025DFPIR, He2025ScaleAdaptive} within a diffusion model to serve as a denoiser. This methodology is applied to a variety of challenging image restoration tasks, including denoising~\cite{Park_2026_CVPR}, deraining~\cite{Dong2025CSUD}, and underwater image reconstruction~\cite{Cheng_2026_WACV}. Our approach exhibits significant improvements in image quality, thereby demonstrating its potential to advance the field of image restoration.

\vspace{1mm}\noindent \textbf{Contributions.} Our main contributions are as follows:
\begin{itemize}
\item We propose TDiR, a transformer-based conditional diffusion model that converts a pretrained PromptIR backbone into a noise-conditioned denoiser, thereby facilitating its application across various restoration tasks.
\item We introduce a lightweight architectural modification in which a dedicated decoder pathway concatenates the degraded input with an embedding of the diffusion timestep. This configuration enables the network to condition its restoration process based on the noise level introduced during the forward pass.
\item We conduct a comprehensive assessment of TDiR across three restoration tasks—underwater image enhancement, denoising, and deraining—utilizing five benchmark datasets. The evaluation compares TDiR against 18 cutting-edge baselines employing four complementary metrics: PSNR, SSIM, UCIQE, and UIQM.
\item We examine the qualitative aspects of the diffusion-based restoration process in comparison with purely discriminative restoration networks, emphasizing its advantages in managing multimodal degradations and acknowledging certain constraints, such as residual patch-based artifacts.
\end{itemize}

\section{Literature Review}
Among the various image restoration tasks~\cite{9113285}, denoising is a vital process that seeks to reduce or eliminate noise within images. Noise may originate in low-light conditions or from errors during image transmission and often appears as graininess, speckles, or random pixel variations that obscure critical visual details. Denoising~\cite{rahman2021structure,anwar2019real} enhances image clarity and quality while preserving essential features, such as edges and textures. Ren~\emph{et~al.}~\cite{Denoise1} proposed a prior-based deep neural network that consistently adapts to various noise levels. In~\cite{Denoise2}, the authors developed a CNN to perform denoising as a step in the image restoration process. Similarly, Zhang~\emph{et~al.}~\cite{Denoise3} used a CNN to address different noise levels by applying a residual learning strategy in their system. The authors of~\cite {Denoise4} implemented a residual learning approach using a deep neural network to effectively extract hierarchical features. Zamir~\emph{et~al.}~\cite{Denoise5} explored the use of generative models and introduced a cycle network for the image denoising task.

Another major challenge in restoration is image de-raining~\cite{Hussain2022Pyramidal}, which involves removing rain streaks and raindrops from images captured outdoors. Rain significantly degrades visibility and image quality, creating noise-like streaks and lowering contrast. These effects make it difficult to accurately identify or analyze objects, especially in applications such as autonomous driving and outdoor surveillance. Deraining improves image clarity by separating rain artifacts and preserving the integrity of the underlying scene. This task is challenging because raindrops and streaks vary in shape, size, and intensity.

In~\cite {rain1}, the authors introduced a transformer-based architecture for reconstructing images disturbed by a rainy effect. A multi-head-tail architecture capable of accomplishing tasks such as deraining and denoising is introduced by Chen~\emph{et~al.}~\cite{9}. Their system requires prior knowledge of the task being fed to the network. On the other hand, the work in~\cite{rain3} presents an architectural search approach, particularly for adverse weather conditions such as deraining, which eliminates the need for prior knowledge. The impressive results in~\cite{rain3} motivated several works to build an inclusive system with higher scalability, in which multiple degradations can be addressed without explicitly notifying the model of the specific type it is dealing with. The authors of~\cite {rain4} developed a three-layer complex network to model the rainy effect in rainy images, along with an additional layer that guides the learning process. Unsupervised learning has also received considerable attention in the literature. Yu~\emph{et~al.}~\cite{rain5} tried to combine model-driven and data-driven approaches for deraining tasks. Their proposed method incorporates unsupervised learning with an optimized deep CNN in a bridge-like architecture.


Another image restoration task is underwater image reconstruction. Underwater images are crucial for human exploration and the development of underwater applications. Underwater photos are typically affected by light scattering, absorption by suspended particles, and color shift. These effects undermine applications that process underwater images, including underwater environment mapping, recognition, and detection. Thus, there is a need for underwater image reconstruction methods that retrieve the underwater scene and remove all artifacts introduced by the water.

Underwater image restoration or reconstruction typically refers to correcting color casts. This problem has been investigated more than other types of degradation. It can be approached as an inverse problem: degradation models of the underwater environment are constructed, their parameters estimated, and the resulting models used to obtain higher-quality images. The most commonly used model is the underwater image formation model (IFM). Such models, especially complex ones, require specialized equipment to measure specific parameters. Researchers typically use priors to estimate model parameters and to restore degraded images~\cite{1, 2, 3, 4}. The Dark Channel Prior~\cite{DCP}, inspired by its use in image dehazing, was adapted for underwater restoration~\cite{UDCP}. However, using priors for parameter estimation limits such reconstruction techniques to cases where these priors hold. For example, with the dark channel prior, results were inferior under non-uniform lighting.

Deep learning-based techniques require large datasets and can be applied without handcrafted priors. Using CNNs showed impressive results for the UIR task~\cite{5, 6, 7, New1Deep}. Inspired by their success in Natural Language Processing, transformers have been investigated for computer vision, and recent models have achieved state-of-the-art results on many high-level vision tasks. However, these architectures are not yet fully addressed for low-level vision tasks such as dehazing, deblurring, denoising, and underwater reconstruction~\cite{8}. Chen~\emph{et~al.}~\cite{9} pre-trained a low-level-vision transformer. The model accommodates different tasks, including super-resolution, denoising, and deraining. This is achieved using multiple pairs of heads and tails, each corresponding to a different task, and a single shared body. A transformer-inspired U-Net architecture was employed for image restoration~\cite{8}. In this work, self-attention is applied to local windows to manage the quadratic complexity of the operation. In addition, several studies incorporate the attention mechanism into underwater image reconstruction~\cite{10, 11, 12, 13}. A U-shaped network with channel-wise and spatial-wise attention is investigated for underwater image reconstruction~\cite{10}. The network is trained using a loss function that considers the RGB, Lab, and LCH color spaces. In~\cite{Tran1}, the authors propose adaptive group attention, which they claim can dynamically select visually complementary channels. Another implementation using transformers can be found in~\cite{Tran2}, where the authors employ a model based on residual dense attention and convolutional blocks. Most of these works use the attention mechanism without employing the full transformer architecture, yet achieve state-of-the-art results on high-level vision tasks.


\begin{figure*}[t]
\centering
\includegraphics[width=0.99\textwidth]{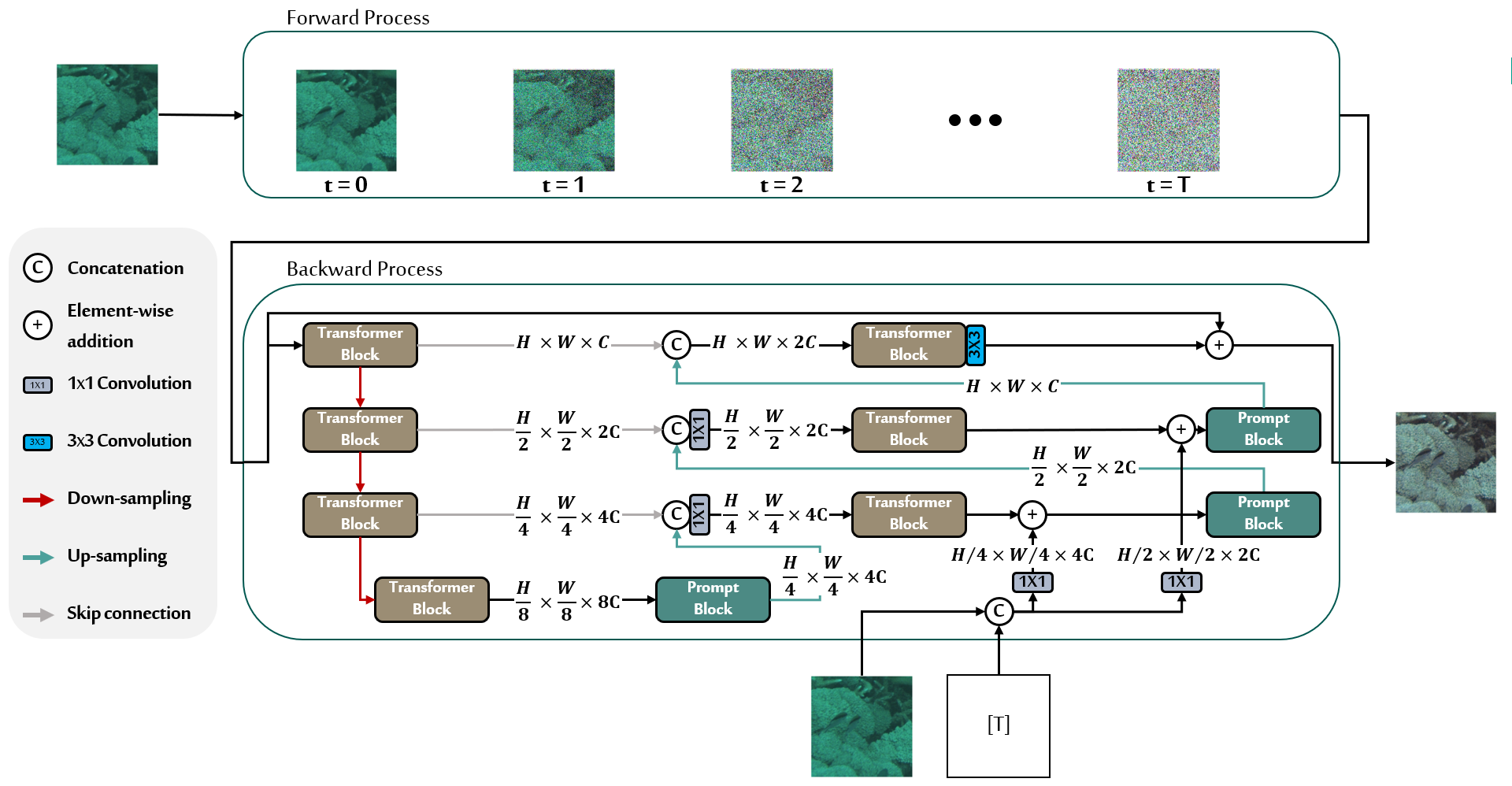}
\caption{Overview of our Approach. The top row shows the forward process. The transformer blocks are used in the Backward process.}
\label{fig:PromprArch}
\end{figure*}

In computer vision, non-generative methods have demonstrated strong performance across a variety of high-level vision tasks. Nonetheless, their effectiveness in addressing challenges in image reconstruction, particularly underwater image reconstruction, does not align with their success on high-level vision tasks. Low-level vision tasks encompass complex forms of image degradation that can significantly hinder these models' ability to achieve accurate, high-quality visual restoration. The degradation observed in underwater imagery varies considerably with lighting conditions, water clarity, and suspended particles, suggesting that a single clear image may originate from multiple degraded sources.

Generative models are particularly effective in this context because they can generate images from inputs that have undergone substantial degradation. Their success primarily arises from their capability to navigate multimodal distributions. This proficiency is imperative for image reconstruction tasks, where the desired outcome may correspond to multiple potential degraded states. Implementations of these methods have yielded promising results~\cite{NewGAN, NewGAN2}. Diffusion models~\cite{DD1, DD2, DD3}, a subclass of generative models, have demonstrated encouraging outcomes in image generation tasks. In contrast to traditional approaches, which may fail to fully capture the diversity of degradation modes present in underwater imagery, diffusion models excel at navigating these complex, multimodal distributions. This enables a more nuanced restoration of underwater images that more closely reflects the real-world variability of these challenging environments.

\section{Methodology}

Diffusion Models consist of a forward diffusion process and a backward denoising stage. The forward operation progressively incorporates Gaussian noise into the input $ y_0 = y$  over $T$ iterations:

\begin{equation}
q(y_{t}|y_{t-1}) = \mathcal{N}(y_{t}; \sqrt{1-\beta_t} y_{t-1}, \beta_t \mathbf{I}),
\end{equation}

\noindent where $\beta_t$ represents the noise scheduler at each timestep. By the final step $T$, $ y_T$ resembles Gaussian noise. We can directly find $y_t$ without going over all the iterations:

\begin{equation}
\alpha_t = 1-\beta_t 
\end{equation}

\begin{equation}
\tilde{\alpha_t}=\prod_{s=1}^t \alpha_s 
\end{equation}

\begin{equation}
q(y_t|y_0)=\mathcal{N}(y_t, \sqrt{\tilde{\alpha_t}} y_0,(1-\tilde{\alpha_t}) I). 
\end{equation}

In the backward denoising process, a denoising network learns to move from the Gaussian distribution to the empirical distribution of the dataset it is trained on.

\vspace{1mm}\noindent\textbf{Transformer Based Denoiser:} Architectures based upon the attention mechanism, such as Transformers, have attained state-of-the-art results across various natural language processing tasks. Given that the transformer architecture eschews recurrence, it displays computational efficiency and scalability. Nevertheless, with respect to the UIR problem and computer vision more generally, convolutional neural network architectures remain the preferred option. Preliminary investigations into the transformer architecture within computer vision initially endeavored to integrate hybrid CNN-self-attention architectures, while alternative approaches contemplated the complete removal of convolutions. Nonetheless, these models either exhibited poor scalability or failed to demonstrate efficacy comparable to that of CNNs.  

In~\cite{PromptIR}, the authors introduce PromptIR, a Transformer-based U-Net architecture designed for image restoration. The model demonstrates encouraging results in addressing various forms of degradation without requiring distinct models for each case. It uses prompt modules to adapt to varying levels of degradation and achieve favorable results on denoising, deraining, and dehazing tasks. In this study, the model from the PromptIR paper is used as the denoiser network.

\vspace{2mm}\noindent\textbf{Objective Function:} For the denoising process during training, given a clear image (Ground Truth), $\boldsymbol{y}$, a degraded version of the image, $\widetilde{\boldsymbol{y}}$, is generated following the forward noising process described in the previous section. Then, a transformer-based U-Net model, $f_\theta$, is trained on the  loss function:

\begin{equation}
   \left\|f_\theta(\widetilde{\boldsymbol{y}}, \gamma)-\boldsymbol{\epsilon}\right\|_1^1,  
\end{equation}

\noindent where $\boldsymbol{\epsilon}$ is the noise affecting the degraded image. The denoiser used in our work is based on PromptIR~\cite{PromptIR}, as shown in Figure~\ref{fig:PromprArch}. We build on this network by adding a path to the decoder that concatenates the input and the noise timestep. Our model shows promising results in handling various degradation types without requiring separate models for each case, using prompt modules to adapt dynamically and achieving strong denoising, deraining, and dehazing performance. 

\begin{figure}[tbp]
\begin{center}
\begin{tabular}[b]{c@{ } c@{ }  c@{ } c}
\multirow{4}{*}{\includegraphics[width=.38\textwidth,valign=t]{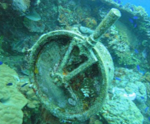}} &  
\includegraphics[trim={0cm 0cm  0.5cm  1cm },clip,width=.19\textwidth,valign=t]{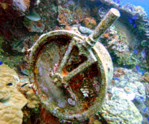}&
\includegraphics[trim={0cm 0cm  0.5cm  1cm },clip,width=.19\textwidth,valign=t]{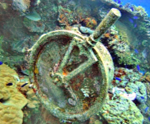}&   
\includegraphics[trim={0cm 0cm  0.5cm  1cm },clip,width=.19\textwidth,valign=t]{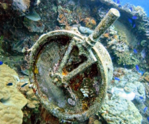}\\
&              & 14.22dB/0.54         & 21.83dB/0.87 \\
& Reference           & Rank1~\cite{Rank1}   & TACL~\cite{TACL}  \\

&
\includegraphics[trim={0cm 0cm  0.5cm  1cm },clip,width=.19\textwidth,valign=t]{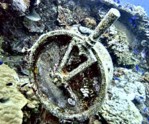}&
\includegraphics[trim={0cm 0cm  0.5cm  1cm },clip,width=.19\textwidth,valign=t]{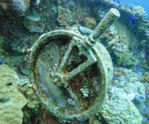}&
\includegraphics[trim={0cm 0cm  0.5cm  1cm },clip,width=.19\textwidth,valign=t]{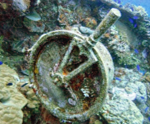}\\
Input Image &15.47dB/0.86     &17.63dB/0.79             &\textbf{17.63dB/0.79}\\
                  &MMLE~\cite{MMLE} &PromptIR~\cite{PromptIR} &TDiR (Ours)\\
    
\end{tabular}
\end{center}
\caption{A real example from UIEB dataset~\cite{UIEB} for comparison of our method against the state-of-the-art algorithms in terms of PSNR and SSIM.}
\label{fig:UIEB1}
\end{figure}















\section{Experimental Settings}

\subsection{Training Setting} 

Since the transformer is reported to be data-hungry, typically attributed to the lack of biases in convolutional neural networks, this issue requires substantial data to learn from. A pre-trained encoder serves as the backbone of the model. Therefore, only part of the decoder is trained in this work, while the encoder parameters remain fixed. The PromptIR model comprises four levels, with prompt blocks inserted after each decoder stage, yielding three prompt blocks. We used an NVIDIA RTX 3090 GPU with 24GB of VRAM and a batch size of 32. The training required approximately four hours across the combined datasets, during which the encoder was kept frozen while only the decoder was optimized, and 128$\times$128 cropped patches were used as input. To augment the training data, we applied random horizontal and vertical flips to the input images. The architecture has fewer parameters to learn than PromptIR—reflecting the additional components that were kept frozen. The baseline was trained with an $\ell_1$ loss using an Adam optimizer ($\beta_1 = 0.9$, $\beta_2 = 0.999$). The noise schedule $\beta_t$ follows a cosine schedule~\cite{nichol2021improved} with 
$T = 1000$ diffusion steps and $\beta_t \in [1\times10^{-4}, 0.02]$.

\begin{figure}[tbp]
\begin{center}
\begin{tabular}[b]{c@{ } c@{ }  c@{ } c}
\multirow{4}{*}{\includegraphics[width=.38\textwidth,valign=t]{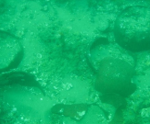}} &  
\includegraphics[trim={0cm 0cm  0.5cm  1cm },clip,width=.19\textwidth,valign=t]{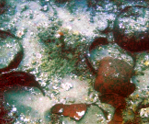}&
\includegraphics[trim={0cm 0cm  0.5cm  1cm },clip,width=.19\textwidth,valign=t]{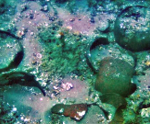}&   
\includegraphics[trim={0cm 0cm  0.5cm  1cm },clip,width=.19\textwidth,valign=t]{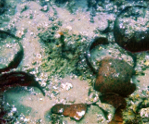}\\
&            & 14.30dB/0.55         & 18.93dB/0.85 \\
& Reference        & Rank1~\cite{Rank1}   & TACL~\cite{TACL}  \\

& \includegraphics[trim={0cm 0cm  0.5cm  1cm },clip,width=.19\textwidth,valign=t]{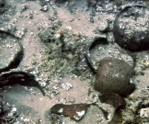}&
\includegraphics[trim={0cm 0cm  0.5cm  1cm },clip,width=.19\textwidth,valign=t]{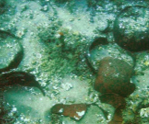}&
\includegraphics[trim={0cm 0cm  0.5cm  1cm },clip,width=.19\textwidth,valign=t]{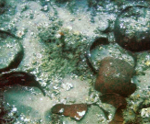}\\
Input Image &15.51dB/0.88     &17.83dB/0.82             &\textbf{22.32dB/0.85}\\
                  &MMLE~\cite{MMLE} &PromptIR~\cite{PromptIR} &TDiR (Ours)\\
    
\end{tabular}
\end{center}
\caption{A challenging example with a strong presence of green color in the image.}
\label{fig:UIEB2}
\end{figure}

\subsection{Evaluation Metrics} 
This study uses four performance metrics to evaluate the quality of the model's output images: two full-reference metrics and two no-reference metrics. For the full-reference measures, we used PSNR and SSIM, while the Underwater Colour Image Quality Evaluation (UCIQE)~\cite{UCIQE} and the Underwater Image Quality Measure (UIQM)~\cite{UIQM} were used to assess the model on no-reference datasets. The UCIQE metric extracts the most informative statistical features from the CIELab color space that characterize underwater image degradation, including color shifts and noise induced by suspended particles. In contrast, the UIQM is a linear superposition of three components: colorfulness, sharpness, and contrast. These two metrics are frequently referenced in the literature for evaluating underwater reconstruction methodologies.

\begin{table*}[tbp]
\centering
\caption{Comparison of our approach against various underwater reconstruction methods on two benchmark datasets. The best results are \textcolor{red}{\textbf{Bold}}, and the second-best results are \textcolor{blue}{\underline{underlined}}.}
\label{tab:Evall}
\begin{tabular}{l|cc||cc}

\toprule
\rowcolor{gray!50}       & \multicolumn{4}{c}{Datasets} \\ \hline
        & \multicolumn{2}{c||}{UIEB~\cite{UIEB}} & \multicolumn{2}{c}{Test-60~\cite{UIEB}}\\

Methods  & PSNR & SSIM & UIQM & UCIQE\\
\midrule

\rowcolor{gray!15}ACDC~\cite{ACDC} & 18.04 & 0.8212 & 1.03 & 0.38 \\

Rank1~\cite{Rank1} & 18.16 & 0.8443 & 1.36 &  0.40 \\

\rowcolor{gray!15}MMLE~\cite{MMLE} & 15.48 & 0.8242 & 1.53 & 0.38 \\

UWCNN~\cite{UWCNN} & 13.43 & 0.6253 & 0.63 & 0.29 \\

\rowcolor{gray!15}WaterNet~\cite{UIEB} & 18.35 & 0.8397 & 1.63 & \textcolor{blue}{\underline{0.41}} \\

Funie~\cite{Funie} & 21.53 & 0.8512 & 1.61 & 0.38 \\

\rowcolor{gray!15}Ucolor~\cite{11} & 20.63 & 0.8495 & 1.63 & 0.34 \\

TACL~\cite{TACL}& 20.41 & 0.8478 & 1.58 & \textcolor{blue}{\underline{0.41}}\\

\rowcolor{gray!15}MetaUE~\cite{MetaUE} & \textcolor{blue}{\underline{22.01}} &\textcolor{blue}{\underline{ 0.8607}} & \textcolor{blue}{\underline{1.65}} & \textcolor{red}{\textbf{0.42}} \\

PromptIR~\cite{PromptIR} & 21.37 & 0.7827 & 1.61 & 0.37 \\

\midrule

\rowcolor{gray!15}TDiR (Ours) & \textcolor{red}{\textbf{22.90}} & \textcolor{red}{\textbf{0.8724}} & \textcolor{red}{\textbf{2.73}} & \textcolor{blue}{\underline{0.41}} \\

\bottomrule

\end{tabular}
\end{table*}
\subsection{Datasets} 

In this study, three distinct image restoration tasks are examined, and for each task, various datasets are used to train and evaluate the model's effectiveness. The aforementioned datasets are detailed in the list below:

\vspace{1mm}\noindent\textbf{Underwater Reconstruction}.  For underwater image reconstruction, the Underwater Image Enhancement Benchmark (UIEB)~\cite{UIEB} dataset is used for training. The UIEB contains 890 underwater images, each accompanied by its corresponding \enquote{reference} image. The reference images were constructed as follows: for each underwater image, several proposed reference images were generated using different underwater image reconstruction methods. Then, some subjects were asked to select the best image among the proposed images that most accurately captured the underwater scene while minimizing color shift. While training on this dataset may mean the model is only as good as the techniques used to produce the references in the UIEB, it is used here because multiple works have published models trained on it. 

\begin{figure}[t]
\begin{center}
\begin{tabular}{c@{ }c@{ }c@{ }c@{ }c@{ }c@{ }c}

\includegraphics[width=.135\textwidth]{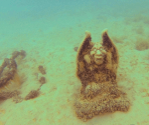}& 
\includegraphics[width=.135\textwidth]{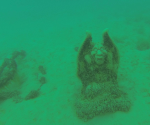}& 
\includegraphics[width=.135\textwidth]{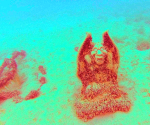}& 
\includegraphics[width=.135\textwidth]{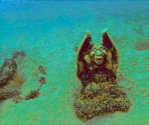}& 
\includegraphics[width=.135\textwidth]{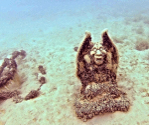}& 
\includegraphics[width=.135\textwidth]{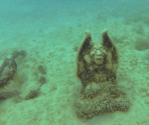}& 
\includegraphics[width=.135\textwidth]{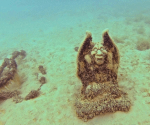}\\
  & 14.53dB & 13.85dB & 18.61dB & 14.96dB & 17.03dB & \textbf{22.39dB} \\

\includegraphics[width=.135\textwidth]{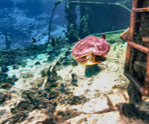}& 
\includegraphics[width=.135\textwidth]{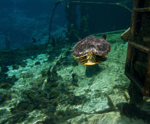}& 
\includegraphics[width=.135\textwidth]{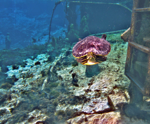}& 
\includegraphics[width=.135\textwidth]{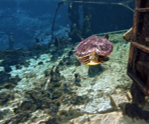}& 
\includegraphics[width=.135\textwidth]{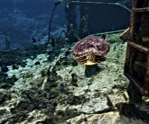}& 
\includegraphics[width=.135\textwidth]{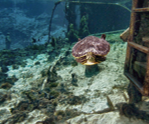}& 
\includegraphics[width=.135\textwidth]{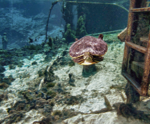}
\\

   & 13.88dB & 14.06dB & 19.26dB  & 14.89dB & 20.77dB  & \textbf{22.46dB} \\
Reference & Noisy   & Rank1 & TACL   & MMLE  & PromptIR & Ours\\


\end{tabular}
\end{center}
    \caption{More examples of comparisons between our model and state-of-the-art on the UIEB dataset, with different backgrounds and noise.}
    \label{fig:UIEB3}
\end{figure}

\vspace{1mm}\noindent\textbf{Denoising}. In order to train the model for the image denoising task, two datasets were utilized: BSD500~\cite{Martin2001BSD} and Waterloo Exploration
Database (WED) \cite{ma2017waterloo}, which contains 4,744 natural images with synthetic Gaussian noise. Images from these datasets were degraded by adding Gaussian noise with $\sigma \in \{15, 25, 50\}$, where $\sigma$ denotes the standard deviation. The testing phase is conducted on the BSD68 Dataset~\cite{roth2009fields}.
    
\vspace{1mm}\noindent\textbf{Deraining}. For the deraining task, the Rain100L dataset~\cite{yang2019joint} was used to train and evaluate the diffusion model, the same dataset used in the PromptIR paper. The dataset comprises 300 clean-rainy image pairs.

\section{Comparisons}

\subsection{Underwater Image Enhancement}
\noindent\textbf{Quantitative Comparison:}
The models are evaluated on two test sets, Test60 and UIEB. For the UIEB dataset, the average PSNR and SSIM are reported in Table~\ref{tab:Evall}, demonstrating that the diffusion model surpasses other techniques across both metrics. TDiR achieves the highest full-reference scores on the UIEB dataset, with 22.90dB PSNR and 0.872 SSIM, exceeding the closest competitor, MetaUE~\cite{MetaUE}, by 0.89dB and 0.011 SSIM points, respectively. Methods based on direct regression, such as WaterNet~\cite{UIEB}, Rank1~\cite{Rank1}, and ACDC~\cite{ACDC}, tend to cluster around 18dB, whereas the least effective baseline, UWCNN~\cite{UWCNN}, underscores the challenges CNN models trained on synthetic data face in generalizing to real-world underwater imagery. Notably, PromptIR~\cite{PromptIR} attains a competitive PSNR of 21.37dB but a low SSIM of 0.7827, suggesting it restores global brightness without fully recovering local structural details. This deficiency is addressed through diffusion-based refinement utilizing an iterative denoising process. Nonetheless, the resulting improvement remains modest, likely attributable to the Diffusion Model's transformer-based denoiser, which processes images in patches. Regarding UCIQE, the model outperforms all methods except the work in~\cite{MetaUE}, which is still not significantly better than ours. On the no-reference Test-60 subset, TDiR achieves a UIQM score of 2.73, exceeding the next-best methods by more than one full unit and reflecting significantly enhanced perceived colorfulness, sharpness, and contrast in its outputs. Regarding UCIQE, TDiR remains competitive with the top methodologies and only slightly trails MetaUE~\cite{MetaUE}, thereby confirming that its CIELab color statistics are comparable to the most advanced existing approaches. 

\noindent\textbf{Qualitative Comparison:} Figures~\ref{fig:UIEB1},~\ref{fig:UIEB2}~and~\ref{fig:UIEB2} provide representative visual comparisons for underwater image enhancement. In Figure~\ref{fig:UIEB1}, the input image is dominated by a strong green color cast typical of shallow tropical waters with suspended particles. Rank1 and TACL produce noticeably different outcomes. TACL recovers more structure but leaves residual color drift, whereas Rank1 tends to oversaturate specific regions. MMLE achieves a high SSIM despite a low PSNR, indicating that its output is structurally similar yet overall inaccurate. PromptIR and TDiR achieve the same accuracy; however, our model demonstrates superior apparent color fidelity by effectively suppressing the green cast. Likewise, Figure~\ref{fig:UIEB2} shows a more extreme case in which the green dominates the entire scene. In this single image, TDiR achieves a gain of 4.49dB over PromptIR, likely because the diffusion model's iterative denoising process more effectively reverses severe, spatially uniform color degradation than a discriminative model trained to map degraded images to their clean counterparts in a single step. Lastly, Figure~\ref{fig:UIEB2} exhibits two rows of images. In the first row (reef scene), visual inspection confirms that TDiR's output closely aligns with the reference in terms of color balance and texture recovery. In the subsequent row (a close-range subject against a flat, featureless backdrop), TDiR shows a smaller visual gap than PromptIR~\cite{PromptIR}; this is consistent with the known difficulty of diffusion models when applied to low-depth scenes. It appears to struggle with images lacking depth, such as scenes presenting a subject against a close background, where color gradients and depth cues that guide the diffusion prior are largely absent. Our model performs well at removing underwater image degradations within challenging scenarios. 

\begin{table*}[tbp]
\centering
\caption{Comparisons on the noisy images against the state of the art on the BSD68~\cite{roth2009fields} dataset. We have used the original models provided by the authors for fair comparisons.}
\label{tab:DiffVs.SOTA_Noising}
\begin{tabular}{l|cc|cc|cc}
\toprule
\rowcolor{gray!50} Methods& \multicolumn{2}{c}{$\sigma$=15} & \multicolumn{2}{c}{$\sigma$=25} & \multicolumn{2}{c}{$\sigma$=50} \\
\midrule

BRDNet~\cite{BRDNet} & 32.26 & 0.898 & 29.76 & 0.836 &  26.34 & 0.836\\
\rowcolor{gray!15}LPNet~\cite{LPNet}  & 26.47 & 0.778 & 24.77 & 0.748 &  21.26 & 0.552\\
FDGAN~\cite{FDGAN}  & 30.25 & 0.910 & 28.81 & 0.868 &  26.43 & 0.776\\
\rowcolor{gray!15}MPRNet~\cite{MPRNet} & 33.54 & 0.927 & 30.89 & 0.880 &  27.56 & 0.779\\
DL~\cite{DL}      & 33.05 & 0.914 & 30.41 & 0.861 &  26.90 & 0.740\\
\rowcolor{gray!15}AirNet~\cite{AirNet} & 33.92 & \textcolor{blue}{\underline{0.933}} & 31.26 & \textcolor{blue}{\underline{0.888}} &  28.00 & \textcolor{blue}{\underline{0.797}}\\


PromptIR~\cite{PromptIR} & \textcolor{blue}{\underline{33.98}} & \textcolor{blue}{\underline{0.933}} & \textcolor{blue}{\underline{31.31}} & \textcolor{blue}{\underline{0.888}} & \textcolor{blue}{\underline{28.06}} & \textcolor{red}{\textbf{0.799}}\\\hline

\rowcolor{gray!15}TDiR (Ours)    & \textcolor{red}{\textbf{34.12}} & \textcolor{red}{\textbf{0.942}} & \textcolor{red}{\textbf{31.53}} & \textcolor{red}{\textbf{0.911}} & \textcolor{red}{\textbf{28.62}} & 0.653\\
\bottomrule
\end{tabular}
\end{table*}


\begin{figure*}[t!]
\begin{center}
\begin{tabular}{c@{ }c@{ }c@{ }c@{ }c}

\includegraphics[width=.19\textwidth]{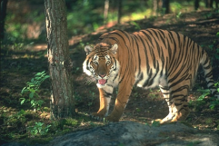}&
\includegraphics[width=.19\textwidth]{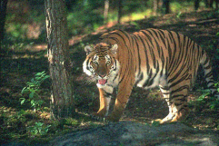}&
\includegraphics[width=.19\textwidth]{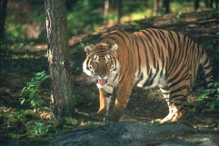}&
\includegraphics[width=.19\textwidth]{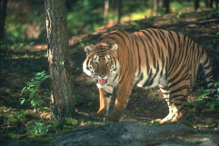}&
\includegraphics[width=.19\textwidth]{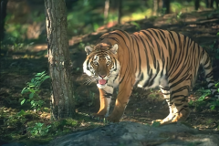}\\
& 23.24/0.63 & 24.43/0.79 & 35.17/0.95 & \textbf{39.17/0.98 }\\

\includegraphics[width=.19\textwidth]{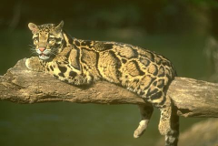}&
\includegraphics[width=.19\textwidth]{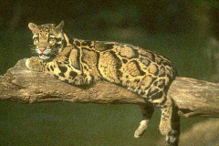}&
\includegraphics[width=.19\textwidth]{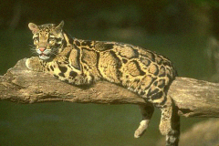}&
\includegraphics[width=.19\textwidth]{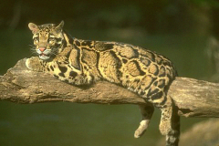}&
\includegraphics[width=.19\textwidth]{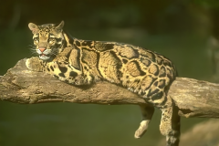}\\
& 21.35/0.58 & 25.81/0.89 & 33.50/0.82 & \textbf{38.9/0.95}\\

\includegraphics[width=.19\textwidth]{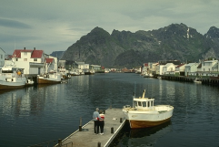}&
\includegraphics[width=.19\textwidth]{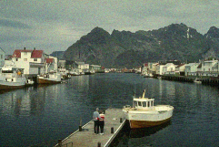}&
\includegraphics[width=.19\textwidth]{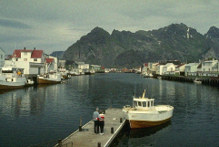}&
\includegraphics[width=.19\textwidth]{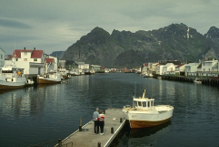}&
\includegraphics[width=.19\textwidth]{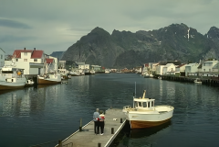}\\
& 22.41/0.69 & 25.01/0.76 & 33.95/0.86 & \textbf{39.85/0.97}
\\

\includegraphics[width=.19\textwidth]{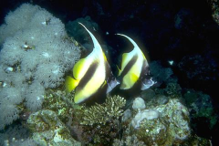}&
\includegraphics[width=.19\textwidth]{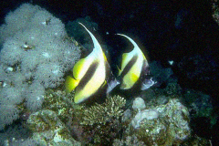}&
\includegraphics[width=.19\textwidth]{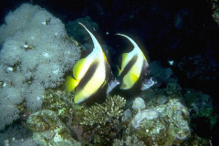}&
\includegraphics[width=.19\textwidth]{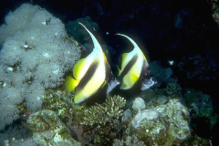}&
\includegraphics[width=.19\textwidth]{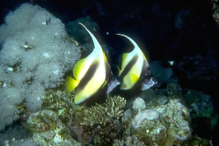}\\
& 24.32/0.7 & 26.45/0.83 & 37.70/0.95 & \textbf{38.95/0.98}\\
Ground Truth & Input & AirNet~\cite{AirNet} & PromptIR~\cite{PromptIR} & TDiR (Ours)\\

\end{tabular}
\end{center}
    \caption{Visual comparisons between our model and the literature on the Denoising task, along with corresponding PSNR/SSIM values.}
    \label{fig:denoise}
    \vspace{-4mm}
\end{figure*}

\subsection{Denoising}

\noindent\textbf{Quantitative Comparison:}
As can be seen in Table~\ref{tab:DiffVs.SOTA_Noising}, our TDiR model outperforms the baseline in all the noise levels tested in the original PromptIR paper. This result is expected, as the way we train the Diffusion Model, where we expose it to different levels of noise, will expose the model to a larger dataset; thus, the model will be better at the task. Another interesting observation is the comparison of the Diffusion Model's results with those of a model trained solely on the denoising task. As seen in Table~\ref{tab:DiffVs.SOTA_Noising}, our model is outperformed by the model trained on a single task; this also makes sense as the denoiser of our Diffusion Model was trained on three different tasks, including denoising, which makes it a more complex training setup, and results in a solution that is further away from an optimal solution for denoising. Surprisingly, our generalist Diffusion Model outperformed the specialist PromptIR model at the $\sigma = 50$ noise level. We have also benchmarked our model against other models available in the literature, as seen in Table~\ref{tab:DiffVs.SOTA_Noising}. Our model outperforms all the benchmark methods. 

\begin{table*}[tbp]
\centering
\caption{TDiR quantitative comparisons are conducted against the current state of the art on the Rain100L~\cite{yang2019joint} deraining dataset. The asterisk (*) indicates that the model was trained on Rain100L~\cite{yang2019joint} to ensure a fair evaluation. }
\label{tab:DiffVs.SOTA_Raining}
\begin{tabular}{l|cc}
\toprule
\rowcolor{gray!50}Methods & PSNR & SSIM\\
\midrule

BRDNet~\cite{BRDNet} & 27.42 & 0.895\\
\rowcolor{gray!15}LPNet~\cite{LPNet} & 24.88 & 0.784\\
FDGAN~\cite{FDGAN}  & 29.89 & 0.933\\
\rowcolor{gray!15}MPRNet~\cite{MPRNet} & 33.57 & 0.954\\
DL~\cite{DL} & 32.62 & 0.931\\
\rowcolor{gray!15}AirNet~\cite{AirNet} & 34.90 & 0.967\\
PromptIR*~\cite{PromptIR}& \textcolor{blue}{\underline{37.03}} & \textcolor{red}{\textbf{0.979}}\\
\rowcolor{gray!15}PromptIR~\cite{PromptIR} & 36.37 & \textcolor{blue}{\underline{0.972}} \\\hline
TDiR (Ours) & \textcolor{red}{\textbf{37.43}} & 0.947\\
\bottomrule
\end{tabular}
\end{table*}

\noindent\textbf{Qualitative Comparison:} Figure~\ref{fig:denoise} presents four image examples with PSNR/SSIM for each method. The figure shows that the diffusion model yields good results when the non-diffusion model's solutions fail. Notice how all other models allow some visible noise to remain in the second and third images, whereas the Diffusion Model solution is more effective at removing it. Overall, our TDiR attains competitive PSNR performance across all noise levels; however, the SSIM slightly decreases at $\sigma = 50$ due to minor patch-based artifacts. The results demonstrate that at moderate-to-high noise levels, TDiR is the only method that yields perceptually clean results. In the first row (tiger), TDiR achieves a 4dB advantage compared to AirNet~\cite{AirNet} and PromptIR~\cite{PromptIR}. In the second row (clouded leopard), the fur texture and spot patterns are visibly better preserved for TDiR than in AirNet~\cite{AirNet} or PromptIR~\cite{PromptIR}. The third and fourth images (harbor scene and underwater shot) follow the same pattern: TDiR consistently achieves the highest accuracy and, on visual inspection, is the only one from which all visible noise has been removed. The mentioned models leave residual grain visible in the flatter regions, i.e., skies, water surfaces, and uniform-coloured backgrounds, whereas the TDiR outputs appear clean throughout, consistent with the known strength of diffusion models in generating perceptually smooth outputs.

\begin{figure*}[t!]
\begin{center}
\begin{tabular}{c@{ }c@{ }c@{ }c@{ }c}

\includegraphics[width=.19\textwidth]{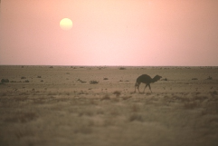}&
\includegraphics[width=.19\textwidth]{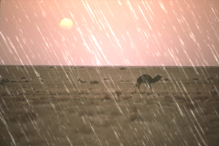}&
\includegraphics[width=.19\textwidth]{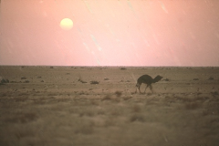}&
\includegraphics[width=.19\textwidth]{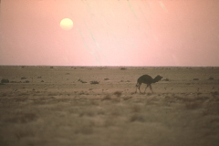}&
\includegraphics[width=.19\textwidth]{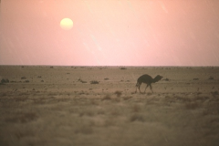}\\
& 22.02dB & 34.83dB & 39.45dB & \textbf{42.67dB} \\

\includegraphics[width=.19\textwidth]{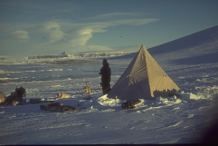}&
\includegraphics[width=.19\textwidth]{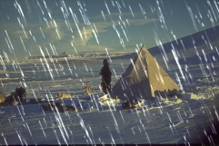}&
\includegraphics[width=.19\textwidth]{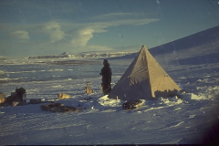}&
\includegraphics[width=.19\textwidth]{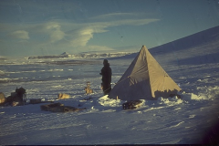}&
\includegraphics[width=.19\textwidth]{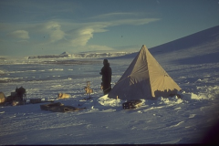}\\
& 20.72dB & 33.00dB & 34.81dB & \textbf{37.61dB}\\

\includegraphics[width=.19\textwidth]{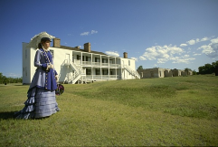}&
\includegraphics[width=.19\textwidth]{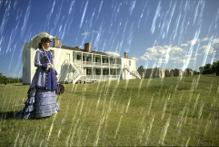}&
\includegraphics[width=.19\textwidth]{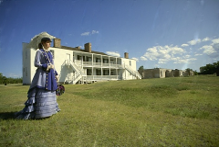}&
\includegraphics[width=.19\textwidth]{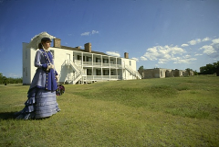}&
\includegraphics[width=.19\textwidth]{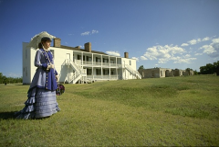}\\
& 20.80dB & 31.21dB & 32.17dB & \textbf{34.45dB}\\

\includegraphics[width=.19\textwidth]{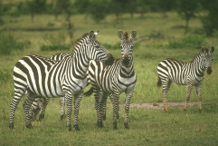}&
\includegraphics[width=.19\textwidth]{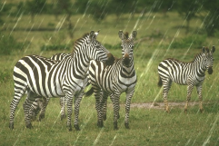}&
\includegraphics[width=.19\textwidth]{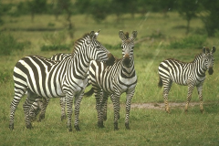}&
\includegraphics[width=.19\textwidth]{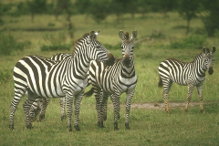}&
\includegraphics[width=.19\textwidth]{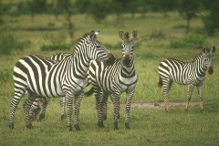}\\
& 23.80dB & 31.09dB & 32.39dB & \textbf{33.89dB}\\
Ground Truth & Input & AirNet~\cite{AirNet} & PromptIR~\cite{PromptIR} & TDiR (Ours)\\

\end{tabular}
\end{center}
    \caption{Visual comparisons between our model and the literature on the deraining task, along with the corresponding PSNR values, are provided. Our model has successfully removed the rain streaks and recovered most of the texture.}
    \label{fig:derain}
    \vspace{-4mm}
\end{figure*}

\subsection{Deraining}
\noindent\textbf{Quantitative Comparison:}
The same principles apply to the deraining task. Notably, the model surpasses both the baseline and the model trained exclusively on deraining, an unexpected outcome. This further demonstrates the effectiveness of Diffusion Models in image restoration tasks. The evaluation of our model against other models documented in the literature has been conducted, as evidenced in Table~\ref{tab:DiffVs.SOTA_Raining}. On the Rain100L~\cite{Rain100L} benchmark, TDiR achieves the highest score among all methods compared, including PromptIR*, which was retrained on Rain100L to ensure a fair comparison. The 0.40dB margin over PromptIR* is particularly significant: it indicates that the diffusion training objective provides a genuine benefit beyond dataset-specific fine-tuning, likely because training within the diffusion framework exposes the denoiser backbone to a wide diversity of noise statistics, effectively treating rain streaks as one degradation mode within a broader learned distribution. Against AirNet~\cite{AirNet}, MPRNet~\cite{MPRNet}, and DL~\cite{DL}, TDiR improves by 2.53dB, 3.86dB, and 4.81dB, respectively, with the gap widening monotonically for older baselines. For SSIM, TDiR trails PromptIR* and PromptIR but exceeds MPRNet and all other methods, again reflecting the patch-boundary artifact pattern that suppresses SSIM regardless of reconstruction accuracy. 

\noindent\textbf{Qualitative Comparison:} Figure~\ref{fig:derain} presents qualitative results for the deraining task. Once again, other techniques fail to capture the entire scene, whereas TDiR consistently eliminates rain streaks in regions of near-uniform color, such as the sky, sand, and grass. Moreover, PromptIR and AirNet leave residual streaks. Fine background textures, such as brickwork and foliage, are better preserved in TDiR's outputs. The largest single-image PSNR gains are observed in high-density rain scenes: 42.67dB versus PromptIR's 39.45dB in the first example of Figure~\ref{fig:derain}, a 3.22dB improvement that is visually confirmed by the complete absence of diagonal streaking in TDiR's reconstruction.

\subsection{Limitations}
Although TDiR demonstrates consistent enhancements across the three aforementioned tasks, certain limitations persist. Due to the transformer-based denoiser's patch-wise image processing, the reconstructed outputs may exhibit conspicuous blocking artifacts at stitch boundaries, which can marginally reduce measured SSIM/PSNR values relative to perceptual quality. Overlapping patches or guided denoising during inference, as employed in previous patch-based diffusion research, may alleviate this issue. The iterative reverse diffusion procedure also requires multiple forward passes per image, making TDiR significantly more costly to infer than single-pass discriminative models. This expense increases proportionally with the number of diffusion steps and could restrict applications requiring real-time performance. Training a single model for denoising, deraining, and underwater reconstruction simplifies deployment; however, our results show that it consistently underperforms compared with specialized single-task models, illustrating a trade-off between versatility and accuracy. Finally, the underwater results indicate that the model has difficulty with scenes at lower depths, such as close subjects set against a flat background, implying a reliance on depth-related cues that are sometimes absent in actual underwater imagery. These issues can be addressed in future work through techniques such as overlapping-patch inference, distillation to fewer diffusion steps, or task-conditioned fine-tuning.

\section{Conclusion}
In this paper, we trained and tested a transformer-based diffusion model on three image restoration tasks: denoising, deraining, and underwater reconstruction. Our model demonstrated improvements over existing methods, achieving the highest PSNR and SSIM for underwater reconstruction on the UIEB dataset, and the best UIQM on the Test-60 dataset, indicating high fidelity and perceptual quality. In denoising, our model outperformed other models across a range of noise levels. Additionally, our model surpassed other state-of-the-art methods on the Rain100L dataset for image deraining. These results illustrate the model's effectiveness in restoring image quality across different scenarios. Although the model performs reliably on denoising, deraining, and underwater enhancement, its ability to generalize varies across degradation types. Certain challenging conditions, such as low-depth underwater scenes or mixed and unclear degradations, can degrade performance, and task-specific fine-tuning may yield additional improvements in such cases. Additionally, we acknowledge practical challenges, including the computational cost of transformer-based diffusion models, sensitivity to hyperparameters, and the risk of patch-based artifacts. 

\bibliographystyle{splncs04}
\bibliography{main-accv}

@ARTICLE{1,  author={Chiang, John Y. and Chen, Ying-Ching},  journal={IEEE Transactions on Image Processing},   title={Underwater Image Enhancement by Wavelength Compensation and Dehazing},   year={2012},  volume={21},  number={4},  pages={1756-1769}}

@InProceedings{Park_2026_CVPR,
    author    = {Park, Junyoung and Oh, Youngjin and Cho, Nam Ik},
    title     = {TM-BSN: Triangular-Masked Blind-Spot Network for Real-World Self-Supervised Image Denoising},
    booktitle = {IEEE/CVF Conference on Computer Vision and Pattern Recognition (CVPR)},
    month     = {June},
    year      = {2026},
    pages     = {29877-29886}
}

@InProceedings{Daroya_2025_ICCV,
    author    = {Daroya, Rangel and Cole, Elijah and Mac Aodha, Oisin and Van Horn, Grant and Maji, Subhransu},
    title     = {WildSAT: Learning Satellite Image Representations from Wildlife Observations},
    booktitle = {IEEE/CVF International Conference on Computer Vision (ICCV)},
    month     = {October},
    year      = {2025},
    pages     = {6143-6154}
}

@InProceedings{Wang_2025_ICCV,
    author    = {Wang, Yuran and Liang, Yingping and Hu, Yutao and Fu, Ying},
    title     = {RobuSTereo: Robust Zero-Shot Stereo Matching under Adverse Weather},
    booktitle = {IEEE/CVF International Conference on Computer Vision (ICCV)},
    month     = {October},
    year      = {2025},
    pages     = {25134-25144}
}

@article{zhang2026out,
  title={Out-of-Sight Embodied Agents: Multimodal Tracking, Sensor Fusion, and Trajectory Forecasting},
  author={Zhang, Haichao and Xu, Yi and Fu, Yun},
  journal={IEEE Transactions on Pattern Analysis and Machine Intelligence},
  year={2026},
  publisher={IEEE}
}

@InProceedings{Cheng_2026_WACV,
    author    = {Cheng, Ching-Heng and Lee, Jen-Wei and Lee, Chia-Ming and Hsu, Chih-Chung},
    title     = {WWE-UIE: A Wavelet \& White Balance Efficient Network for Underwater Image Enhancement},
    booktitle = {IEEE/CVF Winter Conference on Applications of Computer Vision (WACV)},
    month     = {March},
    year      = {2026},
    pages     = {2135-2145}
}

@InProceedings{Qian_2026_CVPR,
    author    = {Qian, Chengcan and Nie, Dong and Chen, Geng and Zhang, Daoqiang and Wen, Xuyun},
    title     = {Simple-ViLMedSAM: Simple Text Prompts Meet Vision-Language Models for Medical Image Segmentation},
    booktitle = {IEEE/CVF Conference on Computer Vision and Pattern Recognition (CVPR)},
    month     = {June},
    year      = {2026},
    pages     = {30042-30052}
}

@article{Anwar2021R2NET,
    title={Attention Based Real Image Restoration},
    author={Saeed Anwar and Nick Barnes and Lars Petersson},
    journal={IEEE Transactions on Neural Networks and Learning Systems},
    year={2021},
    publisher={IEEE}
}

@inproceedings{anwar2015class,
  title={Class-specific image deblurring},
  author={Anwar, Saeed and Phuoc Huynh, Cong and Porikli, Fatih},
  booktitle={IEEE International Conference on Computer Vision},
  pages={495--503},
  year={2015}
}

@article{2,
title = {Automatic Red-Channel underwater image restoration},
journal = {Journal of Visual Communication and Image Representation},
volume = {26},
pages = {132-145},
year = {2015},
issn = {1047-3203},
author = {Adrian Galdran and David Pardo and Artzai Picón and Aitor Alvarez-Gila}
}

@article{yang2019joint,
  title={Joint rain detection and removal from a single image with contextualized deep networks},
  author={Yang, Wenhan and Tan, Robby T and Feng, Jiashi and Guo, Zongming and Yan, Shuicheng and Liu, Jiaying},
  journal={IEEE transactions on pattern analysis and machine intelligence},
  volume={42},
  number={6},
  pages={1377--1393},
  year={2019},
  publisher={IEEE}
}

@article{ma2017waterloo,
	author    = {Ma, Kede and Duanmu, Zhengfang and Wu, Qingbo and Wang, Zhou and Yong, Hongwei and Li, Hongliang and Zhang, Lei}, 
	title     = {{Waterloo Exploration Database}: New Challenges for Image Quality Assessment Models}, 
	journal   = {IEEE Transactions on Image Processing},
	volume    = {26},
	number    = {2},
	pages     = {1004--1016},
	month	  = {Feb.},
	year      = {2017}}

@ARTICLE{3,  author={Drews, Paulo L.J. and Nascimento, Erickson R. and Botelho, Silvia S.C. and Montenegro Campos, Mario Fernando},  journal={IEEE Computer Graphics and Applications},   title={Underwater Depth Estimation and Image Restoration Based on Single Images},   year={2016},  volume={36},  number={2},  pages={24-35}}

@ARTICLE{4,  author={Peng, Yan-Tsung and Cosman, Pamela C.},  journal={IEEE Transactions on Image Processing},   title={Underwater Image Restoration Based on Image Blurriness and Light Absorption},   year={2017},  volume={26},  number={4},  pages={1579-1594}}

@article{5,
title = {Underwater scene prior inspired deep underwater image and video enhancement},
journal = {Pattern Recognition},
volume = {98},
pages = {107038},
year = {2020},
issn = {0031-3203},
author = {Chongyi Li and Saeed Anwar and Fatih Porikli}
}

@ARTICLE{6,  author={Guo, Yecai and Li, Hanyu and Zhuang, Peixian},  journal={IEEE Journal of Oceanic Engineering},   title={Underwater Image Enhancement Using a Multiscale Dense Generative Adversarial Network},   year={2020},  volume={45},  number={3},  pages={862-870}}

@ARTICLE{7,  author={Li, Jie and Skinner, Katherine A. and Eustice, Ryan M. and Johnson-Roberson, Matthew},  journal={IEEE Robotics and Automation Letters},   title={WaterGAN: Unsupervised Generative Network to Enable Real-Time Color Correction of Monocular Underwater Images},   year={2018},  volume={3},  number={1},  pages={387-394}}

@InProceedings{8,
    author    = {Wang, Zhendong and Cun, Xiaodong and Bao, Jianmin and Zhou, Wengang and Liu, Jianzhuang and Li, Houqiang},
    title     = {Uformer: A General U-Shaped Transformer for Image Restoration},
    booktitle = {IEEE/CVF Conference on Computer Vision and Pattern Recognition (CVPR)},
    month     = {June},
    year      = {2022},
    pages     = {17683-17693}
}

@INPROCEEDINGS{9,  author={Chen, Hanting and Wang, Yunhe and Guo, Tianyu and Xu, Chang and Deng, Yiping and Liu, Zhenhua and Ma, Siwei and Xu, Chunjing and Xu, Chao and Gao, Wen},  booktitle={2021 IEEE/CVF Conference on Computer Vision and Pattern Recognition (CVPR)},   title={Pre-Trained Image Processing Transformer},   year={2021},  volume={},  number={},  pages={12294-12305}}

@ARTICLE{10,
  author={Peng, Lintao and Zhu, Chunli and Bian, Liheng},
  journal={IEEE Transactions on Image Processing}, 
  title={U-Shape Transformer for Underwater Image Enhancement}, 
  year={2023},
  volume={32},
  number={},
  pages={3066-3079}
}

@ARTICLE{11,  author={Li, Chongyi and Anwar, Saeed and Hou, Junhui and Cong, Runmin and Guo, Chunle and Ren, Wenqi},  journal={IEEE Transactions on Image Processing},   title={Underwater Image Enhancement via Medium Transmission-Guided Multi-Color Space Embedding},   year={2021},  volume={30},  pages={4985-5000}
}

@INPROCEEDINGS{12,  author={Guo, Zonghui and Guo, Dongsheng and Gu, Zhaorui and Zheng, Haiyong and Zheng, Bing and Wang, Guoyu},  booktitle={OCEANS 2022 - Chennai},   title={Unsupervised Underwater Image Clearness via Transformer},   year={2022},  volume={},  number={},  pages={1-4}
}

@ARTICLE{13,  author={Ren, Tingdi and Xu, Haiyong and Jiang, Gangyi and Yu, Mei and Zhang, Xuan and Wang, Biao and Luo, Ting},  journal={IEEE Transactions on Geoscience and Remote Sensing},   title={Reinforced Swin-Convs Transformer for Simultaneous Underwater Sensing Scene Image Enhancement and Super-resolution},   year={2022},  volume={60},  number={},  pages={1-16}}

@INPROCEEDINGS{DCP,  author={Kaiming He and Jian Sun and Xiaoou Tang},  booktitle={2009 IEEE Conference on Computer Vision and Pattern Recognition},   title={Single image haze removal using dark channel prior},   year={2009},  pages={1956-1963}}

@INPROCEEDINGS{UDCP,  author={Drews Jr, P. and do Nascimento, E. and Moraes, F. and Botelho, S. and Campos, M.},  booktitle={2013 IEEE International Conference on Computer Vision Workshops},   title={Transmission Estimation in Underwater Single Images},   year={2013},  volume={},  number={},  pages={825-830}}

@ARTICLE{Tran1,  author={Huang, Zhixiong and Li, Jinjiang and Hua, Zhen and Fan, Linwei},  journal={IEEE Transactions on Instrumentation and Measurement},   title={Underwater Image Enhancement via Adaptive Group Attention-Based Multiscale Cascade Transformer},   year={2022},  volume={71},  number={},  pages={1-18}}

@Article{Tran2,
AUTHOR = {Sun, Kaichuan and Meng, Fei and Tian, Yubo},
TITLE = {Multi-Level Wavelet-Based Network Embedded with Edge Enhancement Information for Underwater Image Enhancement},
JOURNAL = {Journal of Marine Science and Engineering},
VOLUME = {10},
YEAR = {2022},
NUMBER = {7},
ARTICLE-NUMBER = {884},
ISSN = {2077-1312}
}

@ARTICLE{UIEB,  author={Li, Chongyi and Guo, Chunle and Ren, Wenqi and Cong, Runmin and Hou, Junhui and Kwong, Sam and Tao, Dacheng},  journal={IEEE Transactions on Image Processing},   title={An Underwater Image Enhancement Benchmark Dataset and Beyond},   year={2020},  volume={29},  number={},  pages={4376-4389}}

@article{rahman2021structure,
  title={Structure revealing of low-light images using wavelet transform based on fractional-order denoising and multiscale decomposition},
  author={Rahman, Ziaur and Pu, Yi-Fei and Aamir, Muhammad and Wali, Samad},
  journal={The Visual Computer},
  volume={37},
  number={5},
  pages={865--880},
  year={2021},
  publisher={Springer}
}

@ARTICLE{9113285,
  author={Rahman, Ziaur and Yi-Fei, Pu and Aamir, Muhammad and Wali, Samad and Guan, Yurong},
  journal={IEEE Access}, 
  title={Efficient Image Enhancement Model for Correcting Uneven Illumination Images}, 
  year={2020},
  volume={8},
  number={},
  pages={109038-109053}
}

@ARTICLE{UCIQE,  author={Yang, Miao and Sowmya, Arcot},  journal={IEEE Transactions on Image Processing},   title={An Underwater Color Image Quality Evaluation Metric},   year={2015},  volume={24},  number={12},  pages={6062-6071}}

@ARTICLE{UIQM,  author={Panetta, Karen and Gao, Chen and Agaian, Sos},  journal={IEEE Journal of Oceanic Engineering},   title={Human-Visual-System-Inspired Underwater Image Quality Measures},   year={2016},  volume={41},  number={3},  pages={541-551}}

@article{PromptIR,
  title={Promptir: Prompting for all-in-one image restoration},
  author={Potlapalli, Vaishnav and Zamir, Syed Waqas and Khan, Salman H and Shahbaz Khan, Fahad},
  journal={Advances in Neural Information Processing Systems},
  volume={36},
  pages={71275--71293},
  year={2023}
}

@ARTICLE{New1Deep,
  author={Li, Yinyi and Shen, Liquan and Li, Mengyao and Wang, Zhengyong and Zhuang, Lihao},
  journal={IEEE Transactions on Circuits and Systems for Video Technology}, 
  title={RUIESR: Realistic Underwater Image Enhancement and Super Resolution}, 
  year={2023},
  volume={},
  number={},
  pages={1-1}}

@ARTICLE{NewGAN,
  author={Jiang, Qiuping and Kang, Yaozu and Wang, Zhihua and Ren, Wenqi and Li, Chongyi},
  journal={IEEE Transactions on Multimedia}, 
  title={Perception-Driven Deep Underwater Image Enhancement without Paired Supervision}, 
  year={2023},
  volume={},
  number={},
  pages={1-14}}

@ARTICLE{NewGAN2,
  author={Wang, Haiwen and Yang, Miao and Yin, Ge and Dong, Jinnai},
  journal={IEEE Journal of Oceanic Engineering}, 
  title={Self-Adversarial Generative Adversarial Network for Underwater Image Enhancement}, 
  year={2024},
  volume={49},
  number={1},
  pages={237-248}}

@article{DD1,
  title={Denoising diffusion probabilistic models},
  author={Ho, Jonathan and Jain, Ajay and Abbeel, Pieter},
  journal={Advances in neural information processing systems},
  volume={33},
  pages={6840--6851},
  year={2020}
}

@inproceedings{DD2,
  title={Deep unsupervised learning using nonequilibrium thermodynamics},
  author={Sohl-Dickstein, Jascha and Weiss, Eric and Maheswaranathan, Niru and Ganguli, Surya},
  booktitle={International conference on machine learning},
  pages={2256--2265},
  year={2015},
  organization={pmlr}
}

@article{DD3,
author = {Lu, Siqi and Guan, Fengxu and Zhang, Hanyu and Lai, Haitao},
title = {Underwater image enhancement method based on denoising diffusion probabilistic model},
year = {2023},
publisher = {Academic Press, Inc.},
address = {USA},
volume = {96},
number = {C},
issn = {1047-3203},
journal = {J. Vis. Comun. Image Represent.},
numpages = {10}
}

@ARTICLE{ACDC,
  author={Zhang, Weidong and Wang, Yudong and Li, Chongyi},
  journal={IEEE Journal of Oceanic Engineering}, 
  title={Underwater Image Enhancement by Attenuated Color Channel Correction and Detail Preserved Contrast Enhancement}, 
  year={2022},
  volume={47},
  number={3},
  pages={718-735}
}

@ARTICLE{Rank1,
  author={Liu, Jun and Liu, Ryan Wen and Sun, Jianing and Zeng, Tieyong},
  journal={IEEE Transactions on Pattern Analysis and Machine Intelligence}, 
  title={Rank-One Prior: Real-Time Scene Recovery}, 
  year={2023},
  volume={45},
  number={7},
  pages={8845-8860}}

@ARTICLE{MMLE,
  author={Zhang, Weidong and Zhuang, Peixian and Sun, Hai-Han and Li, Guohou and Kwong, Sam and Li, Chongyi},
  journal={IEEE Transactions on Image Processing}, 
  title={Underwater Image Enhancement via Minimal Color Loss and Locally Adaptive Contrast Enhancement}, 
  year={2022},
  volume={31},
  number={},
  pages={3997-4010}}

@inproceedings{nichol2021improved,
  title={Improved denoising diffusion probabilistic models},
  author={Nichol, Alexander Quinn and Dhariwal, Prafulla},
  booktitle={International conference on machine learning},
  pages={8162--8171},
  year={2021},
  organization={PMLR}
}

@InProceedings{Hussain2022Pyramidal,
    author    = {Liang, Yuanchu and Anwar, Saeed and Liu, Yang},
    title     = {DRT: A Lightweight Single Image Deraining Recursive Transformer},
    booktitle = {IEEE/CVF Conference on Computer Vision and Pattern Recognition (CVPR) Workshops},
    month     = {June},
    year      = {2022},
    pages     = {589-598}
}

@inproceedings{anwar2019real,
  title={Real image denoising with feature attention},
  author={Anwar, Saeed and Barnes, Nick},
  booktitle={IEEE/CVF international conference on computer vision},
  pages={3155--3164},
  year={2019}
}

@article{UWCNN,
title = {Underwater scene prior inspired deep underwater image and video enhancement},
journal = {Pattern Recognition},
volume = {98},
pages = {107038},
year = {2020},
issn = {0031-3203},
author = {Chongyi Li and Saeed Anwar and Fatih Porikli}
}

@ARTICLE{Funie,
  author={Islam, Md Jahidul and Xia, Youya and Sattar, Junaed},
  journal={IEEE Robotics and Automation Letters}, 
  title={Fast Underwater Image Enhancement for Improved Visual Perception}, 
  year={2020},
  volume={5},
  number={2},
  pages={3227-3234}}

@ARTICLE{TACL,
  author={Liu, Risheng and Jiang, Zhiying and Yang, Shuzhou and Fan, Xin},
  journal={IEEE Transactions on Image Processing}, 
  title={Twin Adversarial Contrastive Learning for Underwater Image Enhancement and Beyond}, 
  year={2022},
  volume={31},
  number={},
  pages={4922-4936}
}

@article{MetaUE,
  title={MetaUE: Model-based meta-learning for underwater image enhancement},
  author={Zhang, Zhenwei and Yan, Haorui and Tang, Ke and Duan, Yuping},
  journal={arXiv preprint arXiv:2303.06543},
  year={2023}
}

@inproceedings{Rain100L,
  title={Learning texture transformer network for image super-resolution},
  author={Yang, Fuzhi and Yang, Huan and Fu, Jianlong and Lu, Hongtao and Guo, Baining},
  booktitle={IEEE/CVF conference on computer vision and pattern recognition},
  pages={5791--5800},
  year={2020}
}

@INPROCEEDINGS{Denoise1,
  author={Ren, Chao and He, Xiaohai and Wang, Chuncheng and Zhao, Zhibo},
  booktitle={2021 IEEE/CVF Conference on Computer Vision and Pattern Recognition (CVPR)}, 
  title={Adaptive Consistency Prior based Deep Network for Image Denoising}, 
  year={2021},
  pages={8592-8602}
  }

@INPROCEEDINGS{Denoise2,
  author={Zhang, Kai and Zuo, Wangmeng and Gu, Shuhang and Zhang, Lei},
  booktitle={2017 IEEE Conference on Computer Vision and Pattern Recognition (CVPR)}, 
  title={Learning Deep CNN Denoiser Prior for Image Restoration}, 
  year={2017},
  volume={},
  number={},
  pages={2808-2817}
  }

@article{Denoise3,
author = {Zhang, Kai and Zuo, Wangmeng and Chen, Yunjin and Meng, Deyu and Zhang, Lei},
title = {Beyond a Gaussian Denoiser: Residual Learning of Deep CNN for Image Denoising},
year = {2017},
issue_date = {July 2017},
publisher = {IEEE Press},
volume = {26},
number = {7},
issn = {1057-7149},
journal = {IEEE Transactions on  Image Processing.},
month = {jul},
pages = {3142–3155},
numpages = {14}
}

@article{Denoise4,
  title={Residual Dense Network for Image Restoration},
  author={Zhang, Yulun and Tian, Yapeng and Kong, Yu and Zhong, Bineng and Fu, Yun},
  journal={IEEE Transactions on Pattern Analysis \& Machine Intelligence},
  volume={43},
  number={07},
  pages={2480--2495},
  year={2021},
  publisher={IEEE Computer Society}
}

@INPROCEEDINGS {Denoise5,
author = {S. Zamir and A. Arora and S. Khan and M. Hayat and F. Khan and M. Yang and L. Shao},
booktitle = {2020 IEEE/CVF Conference on Computer Vision and Pattern Recognition (CVPR)},
title = {CycleISP: Real Image Restoration via Improved Data Synthesis},
year = {2020},
volume = {},
issn = {},
pages = {2693-2702},
publisher = {IEEE Computer Society},
address = {Los Alamitos, CA, USA},
month = {jun}
}

@article{roth2009fields,
  title={Fields of experts},
  author={Roth, Stefan and Black, Michael J},
  journal={International Journal of Computer Vision},
  year={2009},
}

@inproceedings{Martin2001BSD,
  author = {D. Martin and C. Fowlkes and D. Tal and J. Malik},
  title = {A Database of Human Segmented Natural Images and its Application to Evaluating Segmentation Algorithms and Measuring Ecological Statistics},
  booktitle = {IEEE/CVF International Conference on Computer Vision (ICCV)},
  year = {2001},
}

@INPROCEEDINGS{rain1,
  author={Jose Valanarasu, Jeya Maria and Yasarla, Rajeev and Patel, Vishal M.},
  booktitle={IEEE/CVF Conference on Computer Vision and Pattern Recognition (CVPR)}, 
  title={TransWeather: Transformer-based Restoration of Images Degraded by Adverse Weather Conditions}, 
  year={2022},
  volume={},
  number={},
}

@INPROCEEDINGS{rain3,
  author={Li, Ruoteng and Tan, Robby T. and Cheong, Loong-Fah},
  booktitle={2020 IEEE/CVF Conference on Computer Vision and Pattern Recognition (CVPR)}, 
  title={All in One Bad Weather Removal Using Architectural Search}, 
  year={2020},
  volume={},
  number={},
  pages={3172-3182}}

@INPROCEEDINGS{rain4,
  author={Zhu, Honghe and Wang, Cong and Zhang, Yajie and Su, Zhixun and Zhao, Guohui},
  booktitle={2020 IEEE International Conference on Multimedia and Expo (ICME)}, 
  title={Physical Model Guided Deep Image Deraining}, 
  year={2020},
  volume={},
  number={},
  pages={1-6}}

@inproceedings{rain5,
author = {Yu, Changfeng and Chang, Yi and Li, Yi and Zhao, Xile and Yan, Luxin},
title = {Unsupervised Image Deraining: Optimization Model Driven Deep CNN},
year = {2021},
isbn = {9781450386517},
publisher = {Association for Computing Machinery},
booktitle = {ACM International Conference on Multimedia},
pages = {2634–2642},
numpages = {9},
location = {Virtual Event, China},
series = {MM '21}
}

@article{BRDNet,
title = {Image denoising using deep CNN with batch renormalization},
journal = {Neural Networks},
volume = {121},
pages = {461-473},
year = {2020},
issn = {0893-6080},
author = {Chunwei Tian and Yong Xu and Wangmeng Zuo}
}

@INPROCEEDINGS{LPNet,
  author={Gao, Hongyun and Tao, Xin and Shen, Xiaoyong and Jia, Jiaya},
  booktitle={2019 IEEE/CVF Conference on Computer Vision and Pattern Recognition (CVPR)}, 
  title={Dynamic Scene Deblurring With Parameter Selective Sharing and Nested Skip Connections}, 
  year={2019},
  volume={},
  number={},
  pages={3843-3851}}

@article{FDGAN,
author = {Dong, Yu and Liu, Yihao and Zhang, He and Chen, Shifeng and Qiao, Yu},
year = {2020},
month = {04},
pages = {10729-10736},
title = {FD-GAN: Generative Adversarial Networks with Fusion-Discriminator for Single Image Dehazing},
volume = {34},
journal = {AAAI Conference on Artificial Intelligence}
}

@INPROCEEDINGS{MPRNet,
  author={Zamir, Syed Waqas and Arora, Aditya and Khan, Salman and Hayat, Munawar and Khan, Fahad Shahbaz and Yang, Ming-Hsuan and Shao, Ling},
  booktitle={2021 IEEE/CVF Conference on Computer Vision and Pattern Recognition (CVPR)}, 
  title={Multi-Stage Progressive Image Restoration}, 
  year={2021},
  volume={},
  number={},
  pages={14816-14826}}

@ARTICLE{DL,
  author={Fan, Qingnan and Chen, Dongdong and Yuan, Lu and Hua, Gang and Yu, Nenghai and Chen, Baoquan},
  journal={IEEE Transactions on Pattern Analysis and Machine Intelligence}, 
  title={A General Decoupled Learning Framework for Parameterized Image Operators}, 
  year={2021},
  volume={43},
  number={1},
  pages={33-47}}

@INPROCEEDINGS{AirNet,
  author={Li, Boyun and Liu, Xiao and Hu, Peng and Wu, Zhongqin and Lv, Jiancheng and Peng, Xi},
  booktitle={2022 IEEE/CVF Conference on Computer Vision and Pattern Recognition (CVPR)}, 
  title={All-In-One Image Restoration for Unknown Corruption}, 
  year={2022},
  volume={},
  number={},
  pages={17431-17441}}

@inproceedings{Dong2025CSUD,
  author    = {Dong, Guanglu and Zheng, Tianheng and Cao, Yuanzhouhan and Qing, Linbo and Ren, Chao},
  title     = {Channel Consistency Prior and Self-Reconstruction Strategy Based Unsupervised Image Deraining},
  booktitle = {IEEE/CVF Conference on Computer Vision and Pattern Recognition (CVPR)},
  year      = {2025}
}

@inproceedings{Tian2025DFPIR,
  author    = {Tian, Xiangpeng and Liao, Xiangyu and Liu, Xiao and Li, Meng and Ren, Chao},
  title     = {Degradation-Aware Feature Perturbation for All-in-One Image Restoration},
  booktitle = {IEEE/CVF Conference on Computer Vision and Pattern Recognition (CVPR)},
  month     = {June},
  year      = {2025}
}

@inproceedings{He2025ScaleAdaptive,
  author    = {He, Xuyi and Quan, Yuhui and Xu, Ruotao and Ji, Hui},
  title     = {A Universal Scale-Adaptive Deformable Transformer for Image Restoration across Diverse Artifacts},
  booktitle = {IEEE/CVF Conference on Computer Vision and Pattern Recognition (CVPR)},
  month     = {June},
  year      = {2025}
}
\end{document}